\def\year{2019}\relax
\documentclass[letterpaper]{article} 
\usepackage{aaai19}  
\usepackage{times}  
\usepackage{helvet}  
\usepackage{courier}  
\usepackage{url}  
\usepackage{graphicx}  
\usepackage{amsfonts}
\usepackage{booktabs}
\usepackage{multirow}
\newcommand{\newcite}[1]{\citeauthor{#1}~\shortcite{#1}}

\newcommand{\enc}{^\text{(enc)}}
\newcommand{\encode}[1]{^\text{(enc,$#1$)}}

\newcommand{\AST}{^\text{(ast)}}
\newcommand{\supscript}[2]{^\text{(#1,$#2$)}}

\newcommand{\yxmodify}[2]{#2}

\newcommand{\yxcomment}[1]{}

\usepackage{amsmath,bm}
\begin{document}
	%
	\title{A Grammar-Based Structural CNN Decoder for Code Generation}
	\author{
		Zeyu Sun$^{\dag}$\ \ Qihao Zhu$^{\dag}$\ \ 
        Lili Mou$^{\ddag}$\ \ 
        Yingfei Xiong\thanks{Yingfei Xiong is the corresponding author. Our code is available at {\tt https://github.com/zysszy/GrammarCNN}}$^{\dag}$\ \ 
        Ge Li$^{\dag}$\ \ 
        Lu Zhang$^{\dag}$\\
		$^{\dag}$Key Laboratory of High Confidence Software Technologies
(Peking University), MoE;\\
Software Institute, Peking University, 100871, P. R. China\\
		\{szy\_, zhuqh, xiongyf, lige, zhanglucs\}@pku.edu.cn\\
  		$^{\ddag}$AdeptMind Research, Toronto, ON, Canada\\
        doublepower.mou@gmail.com
	}
	\maketitle
	\begin{abstract}
		Code generation maps a program description to executable source code in a programming language. Existing approaches mainly rely on a recurrent neural network (RNN) as the decoder. However, we find that a program contains significantly more tokens than a natural language sentence, and thus it may be inappropriate for RNN to capture such a long sequence. In this paper, we propose a grammar-based structural convolutional neural network (CNN) for code generation. Our model generates a program by predicting the grammar rules of the programming language; we design several CNN modules, including the tree-based convolution and pre-order convolution, whose information is further aggregated by dedicated attentive pooling layers. Experimental results on the HearthStone benchmark dataset show that our CNN code generator significantly outperforms the previous state-of-the-art method by 5 percentage points; additional experiments on several semantic parsing tasks demonstrate the robustness of our model. We also conduct in-depth ablation test to better understand each component of our  model.
	\end{abstract}
	
	\section{Introduction}
	\noindent 
	Generating code from natural language description is an important but challenging task in artificial intelligence~\cite{ling2016latent,yin2017syntactic,rabinovich2017abstract,DBLP:journals/chinaf/MeiZ18}. It is beneficial to various applications. For example, a programmer would like to ``open the file, F1'' in Python, but does not know how to implement it in the programming language. Hopefully, he or she can obtain the target code ``\texttt{f = open('F1', 'r')}'' by code generation. 
	
	With the prosperity of deep learning, the encoder-decoder framework becomes a prevailing approach to sequence generation. In particular, recurrent neural networks (RNNs) typically serve as the encoder and decoder; such architecture is also known as a sequence-to-sequence (Seq2Seq) model~\cite{sutskever2014sequence}. When applied to code generation, it takes the program description as the input sequence and generates the desired code as the output sequence~\cite{ling2016latent}.

	\begin{figure}
		\centering 
		\includegraphics[width=.7\linewidth]{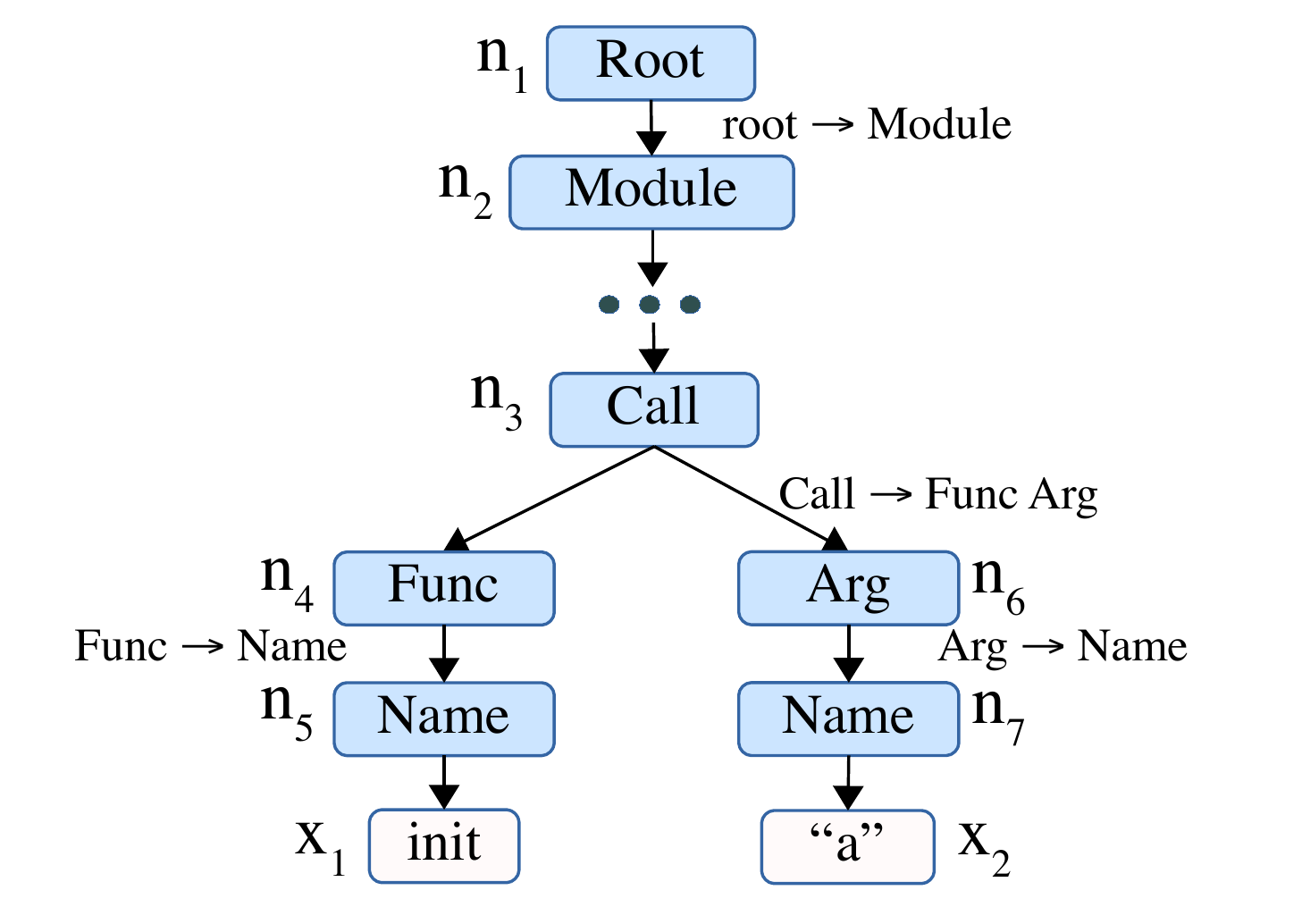}
		\caption{The abstract syntax tree (AST) of code: \texttt{init(a)}.}
        \label{fig:ast_eg}
	\end{figure}
    
	It has been pointed out that programs contain rich structural information, which is important to program modeling~\cite{rabinovich2017abstract,yin2017syntactic}. However, traditional Seq2Seq neural networks do not explicitly model program structures.  Figure~\ref{fig:ast_eg} shows an example of a Python abstract syntax tree (AST), where the two nodes $\mathrm{n}_3$ and $\mathrm{n}_6$ should have interacted intensively as parent-child nodes, but are far away from each other if the tree is pre-order traversed to a sequence. This brings difficulties to Seq2Seq models.

	To address this problem, \newcite{dong2016language} proposed an approach that generates code along the abstract syntax tree (AST) of a program, but their generation is still in the token level. More recent work generates programs by predicting the grammar rule or rewriting rule to apply at each step~\cite{DBLP:conf/icse/XiongWFZ18,yin2017syntactic,rabinovich2017abstract}; thus, the generated programs are guaranteed to be syntactically correct. When neural network is used in those approaches, an RNN is used to capture the \textit{autoregressiveness}\footnote{By ``autoregressive,'' we mean that, during decoding, a step of prediction is dependent on previous decoded steps.} of predictions within the decoder.
	\begin{table*}[!t]
		\centering
        \resizebox{\textwidth}{!}{
		\footnotesize  
		\begin{tabular}{l|p{15cm}}
			\toprule
            \textbf{Rule} & \textbf{Explanation}\\
			\midrule
			\tt stmt $\rightarrow$ If | For | ... & A statement (\texttt{stmt}) could be an \texttt{If}-block, a \texttt{For}-block, and many others. They are different rules, and the network predicts the most appropriate rule to apply at a time step.\\ \midrule
           \tt If $\rightarrow$ expr  stmt*  stmt* &
           An \texttt{If}-block starts with a testing expression (\texttt{expr}). If it is true, the first statement list is executed, or otherwise, the second statement list is executed. In the official python grammar, a rule may generate a list of tokens (e.g., \texttt{stmt*}) with an arbitrary length. We examine the training samples and treat each length as a separate rule to predict.\\ 
			\bottomrule   
		\end{tabular}}
		\caption{Examples of python grammar rules.}
		\label{table:rules}
	\end{table*}
    
	In the deep learning community, researchers are showing growing interest in using the convolutional neural network (CNN) as the decoder~\cite{gehring2017convolutional,chaturvedi2018cnn}, because of its efficiency and easiness of training. We further observe that a program is much larger than a natural language sentence and that RNNs---even with long short-term memory \cite[LSTM]{hochreiter1997long} units---suffer from the long dependency problem~\cite{bengio1994learning}. CNNs, on the contrary, are able to capture features effectively at different regions by sliding windows.
	
	To this end, we propose a grammar-based structural CNN for code generation. Our model generates code by grammar rules of construction in AST, e.g., \texttt{If} $\rightarrow$ \texttt{expr stmt* stmt*}, {following the framework in our previous work~\cite{DBLP:conf/icse/XiongWFZ18}.} Since the sequence of child nodes is generated by one step of prediction, it enables more compact prediction than the token-by-token generation. In other words, our model predicts the sequence of grammar rules, which eventually form an entire program. 
	
	In our approach, the prediction of a grammar rule is mainly based on three types of information: the source sequence that specifies the program to be generated, the previously predicted grammar rules, and the partial AST that has been generated. Here, the former one is the input to the encoder. The latter two enable the autoregressiveness of the decoder, and as usual, the decoder is also conditioned on the encoder.

	We design several distinct components for the structural CNN, suited to program generation: (1) We first adopt an idea of tree-based convolution that applies sliding windows on the AST structures~\cite{mou2016convolutional}. Then we design another CNN module to the pre-order traversal of nodes in the partial AST. These two types of CNNs capture neighboring information not only in the sequence but also in the tree structure.
    (2) To enhance ``autoregressiveness,'' we apply another CNN module to the ancestors of the node to be generated, and thus the network is aware of where to generate at a certain step.
	(3) We design a dedicated attentive pooling mechanism that aggregates CNN features interacting with different neural modules. In particular, we find it useful to consider the scope names (e.g., function and method names) during code generation, and use such information as the controller of several attentive pooling layers.

We conducted experiments on an established benchmark dataset, HearthStone, for python code generation~\cite{ling2016latent}. Experimental results show that our CNN-based code generator outperforms previous RNN approaches to a large extent. We further evaluate our approach on two semantic parsing tasks, where the target programs are shorter than HearthStone; our approach also achieves comparable results to previous state-of-the-art methods, indicating the robustness of our method. We  conducted extensive ablation tests, showing that our design of grammar-based structural CNN is better than applying CNN in a na\"ive fashion.

	To the best of our knowledge, we are the first to successfully apply CNN decoders for code generation.

	\section{The Proposed Model}
\begin{figure}[!t]
  		\includegraphics[width=\linewidth]{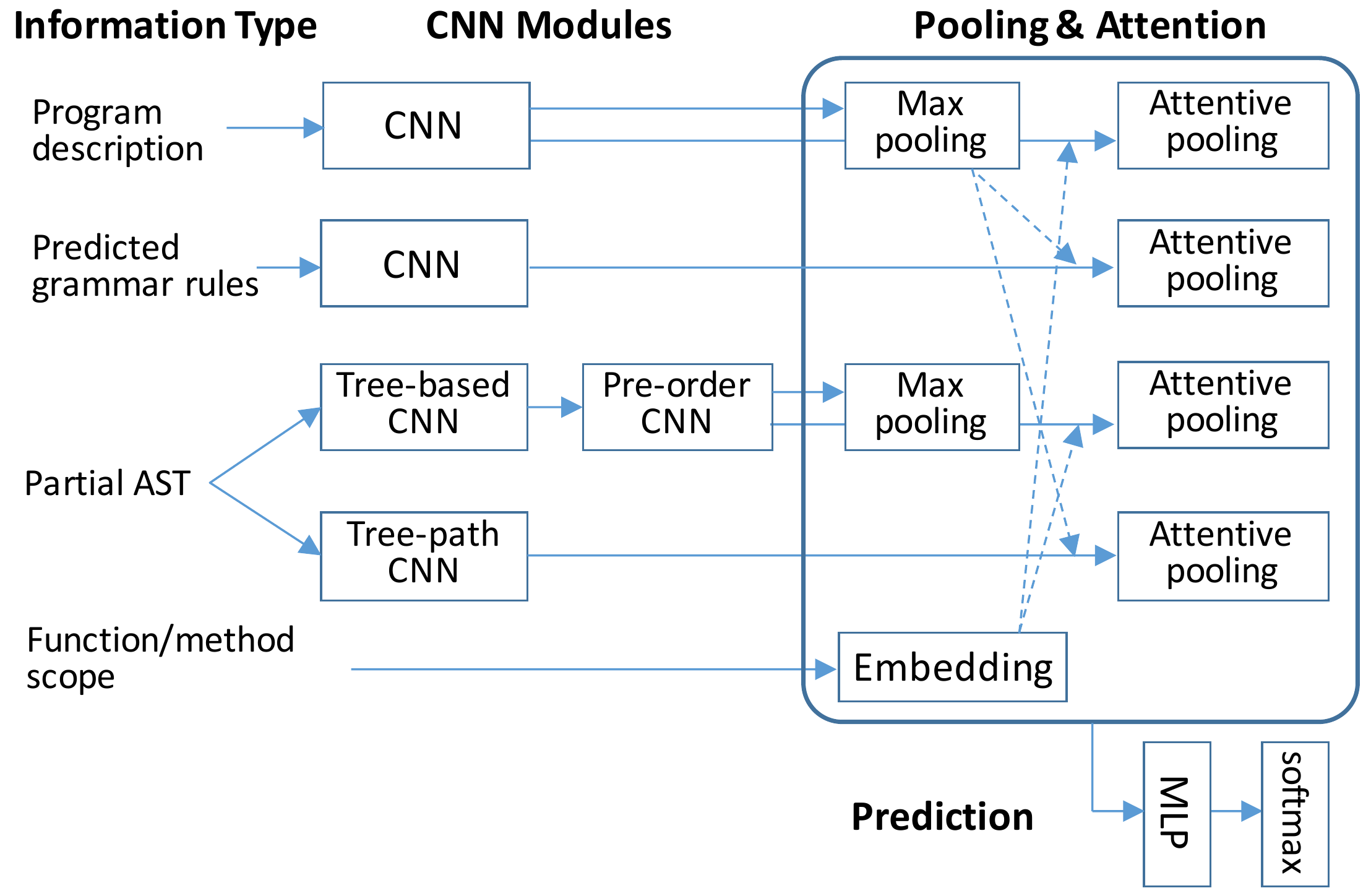}   
		\caption{Overview of our model. The dashed arrows indicate attention controllers.}
		\label{fig:nn}
	\end{figure}

  Figure~\ref{fig:nn} shows the overall structure of our network. We will first describe the process of grammar-based code generation, and then introduce each module in detail.
  
	\subsection{Grammar-Based Code Generation}
	For an input of program description, our task is to generate a piece of executable code that complies with the description.
	In traditional Seq2Seq models, a program can be represented as a sequence of tokens $\mathrm{x}_1, \mathrm{x}_2, \cdots, \mathrm{x}_T$, and these tokens are generated in sequence.

	Alternatively, a valid program can be represented by an abstract syntax tree (AST) in Figure~\ref{fig:ast_eg}. Leaf nodes are the terminal symbols, denoted as $\mathrm{x}_1, \cdots, \mathrm{x}_T$. Non-leaf nodes are non-terminal symbols $\mathrm{n}_1, \cdots, \mathrm{n}_{N}$, each representing an abstract component of the program (e.g., an \texttt{If}-block). Moreover, child nodes $\mathrm n_1,\cdots, \mathrm n_k$, stemming from their parent node $\mathrm p$, are obtained by applying some grammar rule $\rm r$, denoted as $\mathrm p \overset{\mathrm r}{\rightarrow} \mathrm n_1\cdots\mathrm n_k$. In our work, leaf nodes of different user-defined variables are treated as separate grammar rules by examining the training set. Table~\ref{table:rules} illustrates several Python rules and their meanings.\footnote{Full list available at {https://docs.python.org/2/library/ast.html}}
	
	\newcite{dong2016language} propose to generate an executable command by following the AST, but they predict child nodes $n_1\cdots n_k$ one at a time with an RNN. In our study, we follow more recent work~\cite{rabinovich2017abstract,yin2017syntactic,DBLP:conf/icse/XiongWFZ18}, predicting the rules $\mathrm r_1, \mathrm r_2,\cdots,\mathrm r_M$ that generate the program. We traverse the tree in depth-first pre-order, and for the first encountered non-terminal symbol, we predict what rule should be used to expand it. In other words, the probability of a program is decomposed as 
	\begin{equation}
	p(\text{program})=\prod\nolimits_{n=1}^M p(\mathrm r_n|\mathrm r_1\cdots,\mathrm r_{n-1})
	\label{eq:program}
	\end{equation}
	Although a typical programming language contains more grammar rules than distinct AST nodes, grammar-based generation is more compact because the child nodes $\mathrm c_1, \cdots, \mathrm c_k$ become in place by a single prediction of the rule $\mathrm p \overset{\mathrm r}{\rightarrow} \mathrm c_1\cdots\mathrm c_k$. Moreover, the generated program is guaranteed to be syntactically correct.
	
	In the rest of this section, we describe our CNN encoder-decoder model for the prediction of grammar rules.

	\subsection{CNN for the Input}
	The input of our model is a piece of description that specifies the program to be generated. For code generation of a card in HearthStone, the input is semi-structured data, containing the card's name, properties, and descriptions, illustrated in Figure~\ref{fig:hearth}a. For other tasks like semantic parsing~\cite{zettlemoyer2012learning}, the input could be a natural language sentence.
	
	We tokenize the input, and obtain a sequence of tokens $\mathrm x\enc_1, \cdots, \mathrm x\enc_I$, where $I$ is the length of the input. The tokens are represented as real-valued vectors $\bm x\enc_1, \cdots,\bm x\enc_I$, known as \textit{embeddings}. 
	
	Then, a set of convolutional layers are applied and extract features $\bm y\encode{L}_1, \cdots, \bm y\encode{L}_I$. In particular, we adopt shortcut connections every other layer parallel to linear transformation before the activation function, as in ResNet \cite{he2016deep}. This helps the training of a deep neural network.
	
	Formally, the extracted features $\bm y_i\encode{l}$ are computed by
	\begin{equation}
	\begin{split}
	\bm y_i\encode{l} = &\operatorname{ReLU}\big( c\encode{l} \cdot \bm y\encode{l\!-\!2}_{i-1} \\&+ W\encode{l}[\bm y\encode{l\!-\!1}_{\lceil {i-s} \rceil};\cdots;\bm y\encode{l\!-\!1}_{\lceil {i+s} \rceil}]\big) \label{eq:CNN}
	\end{split}
	\end{equation}
	where $W\encode{l}$ are the convolution weights for the encoder CNN, $s$ is computed by $s = (k - 1)/2$, $k$ is the window size (set to 2 in our experiment), and $l=1,\cdots, L$ indicates the layer in the deep CNN. In particular, $\bm y_i\encode{0}$ is the input embedding $\bm x_i\enc$. $c\encode{l}=1$ for even layers and $0$ for odd layers, indicating whether the shortcut connection exists for this layer. For the first and last several words, we perform zero padding.

	\subsection{CNN for Predicted Rules}
	Since the probability of a program is decomposed by grammar rules (Equation~\ref{eq:program}), we keep track of all previously predicted rules, and build a deep neural network to extract such information. 
	
	Let $\mathrm r_1, \cdots,\mathrm r_{n-1}$ be the previously predicted rules. We embed them as real-valued vectors, $\bm r_1,\cdots, \bm r_{n-1}$, where the embeddings are randomly initialized and learned by back-propagation.
	
	We apply a deep CNN module with shortcut connections  to rule embeddings $\bm r_1,\cdots, \bm r_{n-1}$, extracting features $\bm y_1\supscript{rule}{L}, \cdots, \bm y_{n-1}\supscript{rule}{L}$. The computation is the same as Equation~\ref{eq:CNN}, but with different weight parameters. Details are not repeated here.

The predicted grammar rules fully specify the generated (partial) program in a compact fashion, which is beneficial for accurate code generation. 

However, it is improper to feed the decoder with only predicted rules for autoregressiveness, as they do not provide a concrete/pictorial view of the program due to the compactness. To alleviate this problem, we enhance the decoder with the partial AST, described below.

\subsection{CNN for Partial AST}
	
	An abstract syntax tree (AST) is a tree-structured representation of a program, where a rule $\mathrm p\overset{\mathrm r}{\rightarrow} \mathrm n_1\cdots\mathrm n_k$ is expanded as parent-child edges of $\mathrm p$ and $\mathrm n_1,\cdots,\mathrm n_k$. 
	
	We design a deep CNN module to capture AST's structural information. It contains tree-based convolutional layers, pre-order traversal convolutional layers, as well as a tree-path CNN submodule to inform the network of where the next grammar rule is applied.
	
	\begin{figure}
		\centering  
		\includegraphics[width=\linewidth]{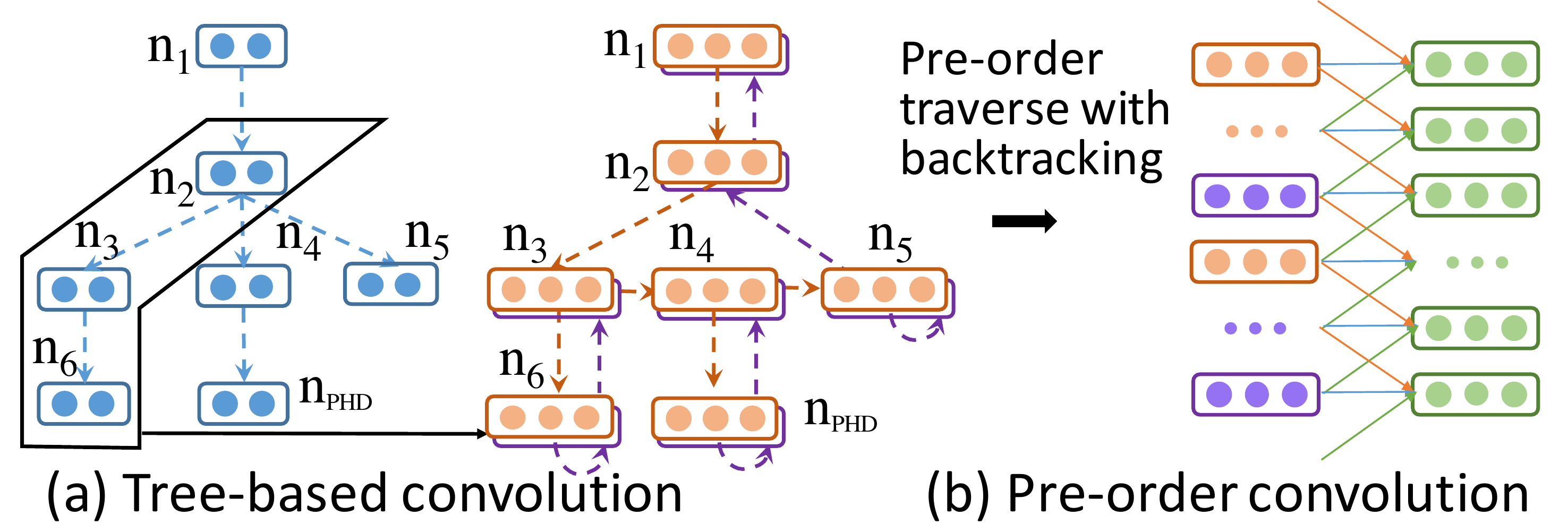}
		\caption{CNN for the partial AST. The dashed arrows are not a part of the neural network, but indicate topologically neighboring information. In particular, the purple up arrows are backtracking traces.}  
		\label{fig:tree_conv}
	\end{figure}
	
	\subsubsection{Tree-Based CNN.}
	We first apply a tree-based CNN to the partial AST, similar to \citeauthor{mou2016convolutional}~\shortcite{mou2016convolutional,tbcnn_book}. The main intuition  is to design a local feature detector of a fixed depth, sliding over a tree to extract structural features, shown in Figure~\ref{fig:tree_conv}a. 
	
The input of tree-based CNN is the partial AST that has been generated, each node represented by an embedding. We also put a placeholder node ($\mathrm n_\text{PHD}$ illustrated in Figure~\ref{fig:tree_conv}) to indicate where the next grammar rule is applied.

	Suppose a node $\mathrm n_i$ has a parent node $\mathrm p_i$ and a grandparent node $\mathrm g_i$, and their vector representations are $\bm n_i, \bm p_i,$ and $\bm g_i$, respectively. Then the tree-based CNN extracts features $\bm y\AST_1, \cdots, \bm y\AST_{n-1}$, computed by
	\begin{equation}
	\bm y\AST_i = \operatorname{ReLU}(W\AST [\bm n_i; \bm p_i; \bm g_i])
	\end{equation}
	where $W\AST$ is the weight of the tree-based convolution kernel. We pad a special token for the nodes in the top two layers who do not have a parent and/or grandparent.

	Note that our tree-based convolution slightly differs from \newcite{mou2016convolutional} in that we have a deeper window but do not consider sibling information.
	This is because our grammar-based generation obtains all siblings at a time by applying a certain rule, and hence, the siblings are less important than ancestors. The complexity of \newcite{mou2016convolutional} unfortunately grows exponentially with  depth, and is less tractable,  whereas our tree-based CNN variant grows linearly. In terms of the convolution computation, we follow \newcite{mou2016convolutional} and adopt a perceptron-like interaction. Deep tree-based convolution and shortcut connections as in ResNet could be explored as future work.
	
	\subsubsection{Pre-Order Traversal CNN.}	After obtaining a set of vectors extracted by tree-based CNN, we apply a pre-order traversal convolution with $\bm y\AST$ being input (Figure~\ref{fig:tree_conv}b).  That is, the AST nodes are organized in a sequence by pre-order traversing.
    
    It can be shown that a simple pre-order traverse is not invertible to the tree structure, i.e., different tree structures could yield a same sequence. To address this problem, we keep track of backtracking traces during pre-order traversing. For example, the AST in Figure~\ref{fig:tree_conv} yields the sequence $\mathrm n_1, \mathrm n_2, \mathrm n_3, \mathrm n_6, \mathrm n_6^\text{(bt)}, \mathrm n_3^\text{(bt)}, \mathrm n_4, \mathrm n_4^\text{(bt)},\mathrm n_\text{PHD}, \mathrm n_\text{PHD}^\text{(bt)}, \mathrm n_5, \mathrm n_5^\text{(bt)},\mathrm n_2^\text{(bt)},$ $\mathrm n_1^\text{(bt)}$. Their vector representations (including backtracking nodes and the placeholder node) are predicted by tree-based convolution. Then a deep CNN as in Equation~\ref{eq:CNN} is applied to extract features $\bm y_1\supscript{tree}{L}, \cdots, \bm y_{2S}\supscript{tree}{L}$, where $L$ is the number of CNN layers. $T$ is the number of nodes in the AST, and pre-order traverse with backtracking yields $2S$ input units.

	It should be noted that the tree-based CNN and the pre-order traversal CNN capture different information. Pre-order traverse yields an order that addresses sequential neighborhood of an AST node during generation, whereas tree-based convolution enables information fusion for nodes that are structurally neighboring. In Figure~\ref{fig:tree_conv}, for example, node $\mathrm n_4$ is a child node of $\mathrm n_2$. However, after the generation of some other part of the program (namely, $\mathrm n_3$ and $\mathrm n_6$), the nodes $\mathrm n_2$ and $\mathrm n_4$ are no longer close to each other. Tree-based convolution directly builds a feature extractor for a node and its ancestors to enable their interaction. Therefore, we believe these two types of CNNs are complementary to each other.

	\subsubsection{Tree-Path CNN.} Should we only consider the above CNNs, it would be hard for the model to tell the position where the next grammar rule is applied. For example, the tree-based CNN and the pre-order traversal CNN would yield very similar features if we expand $\mathrm n_4$ or $\mathrm n_5$ in Figure~\ref{fig:tree_conv}, despite the placeholder we introduce for pre-order CNN.
	
	Technically speaking, if we follow leftmost derivation, then where the next rule is applied is unambiguous. But such clue is too implicit and should be modeled more explicitly.

	We thus extract the path from the root to the node to expand. For example, if we are about to expand $\mathrm n_4$, the path should be $\mathrm n_1, \mathrm n_2, \mathrm n_4$.
    Then a set of convolutional layers extract features ${\bm y}^\text{(path,$L$)}_{1},\cdots,\bm{y}^\text{(path,$L$)}_{J}$, also computed as Equation~\ref{eq:CNN}. ($J$ is the number of nodes in the path, and $L$ is the number of CNN layers.) We call this \textit{tree-path convolution}.

	\subsection{Pooling and Attention Mechanisms}
	
	CNNs extract a set of features with the same size or shape as input. To facilitate softmax prediction for code generation, we need to aggregate information into one or a few fixed-size vectors, regardless of the input size.
	
	Traditionally, people use max pooling for CNNs~\cite{krizhevsky2012imagenet} as well as tree-based CNNs~\cite{mou2016convolutional,tbcnn_book}. However, this makes the underlying CNN modules separate and unable to communicate during information aggregation.
	
	Therefore, we incorporate attention mechanisms for CNN pooling, similar to \newcite{yu2018qanet}. Essentially, an attention mechanism computes a weighted sum of a set of candidate features (extracted by CNN), where the weights are computed by a controlling vector (for example, a max pooling vector for another CNN module).
	
	Formally, given a controlling vector $\bm c$ and a set of candidate convolutional features $\bm y_1,\cdots,\bm y_D$ extracted by a CNN module ($D$ is the number of feature vectors), we compute attention logit by\\[-.7cm]
    
	\begin{equation}\label{eqn:control}
	\widetilde\alpha_i =\bm y_i^\top  W^\text{(att)}  \bm c
	\end{equation}
	where $W^\text{(att)}$ is a trainable matrix, inspired by metric learning.
	Then the attention weight $\alpha_i$ for the node $i$ is
	\begin{equation}
	\alpha_i= \frac{\exp\{\widetilde\alpha_i\}}{\sum_{j=1}^D \exp\{\widetilde\alpha_j\}}
	\end{equation}
	Finally, the attentive pooling yields a vector $\bm y^\text{(att)}$ by
	\begin{equation}
	\bm y^\text{(att)} = \sum\nolimits_{i=1}^D \alpha_i \bm y_i
	\end{equation}
	
	To apply such an attentive pooling layer to our underlying CNNs, we consider several key information as the controlling vector. (1) The input description specifies the program to be generated, and we use it to control the grammar rule CNN and the tree-path CNN. In particular, we apply a max pooling layer to aggregate input CNN features as a fixed-size controlling vector, which is used to compute the attention weights for tree-path CNN and the CNN for predicted grammar rules. (2) We note that a \textit{scope name} (namely, a function name or a method name) provides illuminating information about its descendants. Such information is not captured by AST node types, and thus we embed the scope name as a vector and use it to control the pre-order traversal CNN and the CNN for the input. It should be noted that if the current program snippet is under two or more scopes (a function and a method), we only consider the nearest scope as the controlling vector. If the code snippet does not belong to any function or class, then the scope embedding is set to a zero vector.

	In addition to the attentive pooling for the tree-based convolution, it is useful to apply another max pooling layer to the pre-order traversal CNN features. Our empirical finding is that the controlling scope embedding makes the attention too peaked at the corresponding AST node, and that aggregated information is not sufficient. Another max pooling layer could preserve more information regardless of the controlling vector.
	
	We would also like to point out that there are different choices of designing the attention mechanism and its controlling connections in a deep neural network with multiple modules. For example, we might as well use the CNN for the input to control all other modules, following the spirit of attention in the encoder-decoder framework. However, our pilot experiment shows that such design yields a worse performance, and thus we adopt the current architecture.
	
	\subsection{Training and Inference} 
	We concatenate all max pooling and attentive pooling layers. They are fed to a two-layer perceptron, where the last layer has a softmax activation function for predicting the next grammar rule, given by
	\begin{equation}
	p(\mathrm r_i|\cdot)= \frac{\exp\{h^\text{(MLP)}_i\}}{\sum_{j=1}^R \exp\{h^\text{(MLP)}_j\}}
	\end{equation}
	where $h_i^\text{(MLP)}$ is the input logit of softmax, and $R$ is the number of candidate grammar rules. 
	
	\begin{figure}[!t]
		(a)\\
		\begin{minipage}{0.99\linewidth}
        	\centering
            \includegraphics[width=0.99\linewidth]{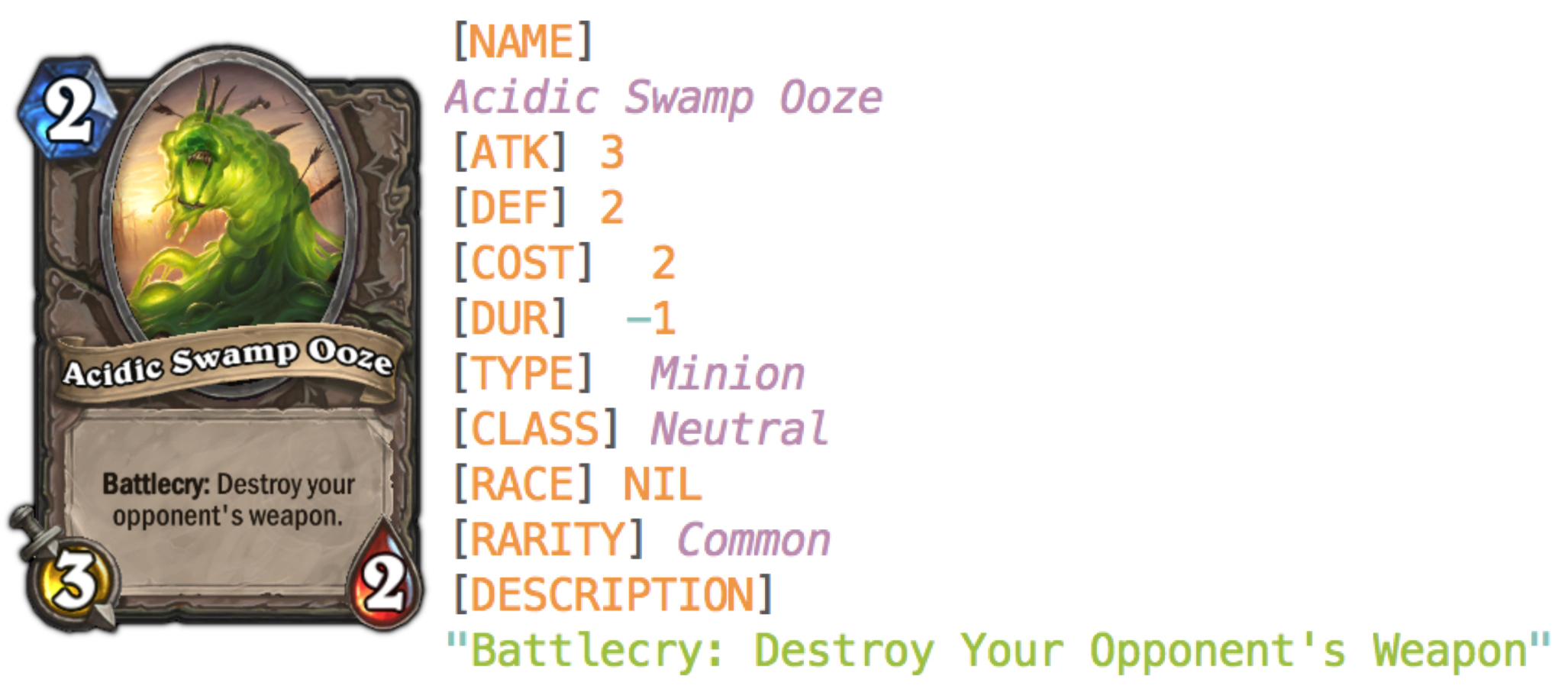}
		\end{minipage}
		(b)\\
		\noindent\includegraphics[width=\linewidth]{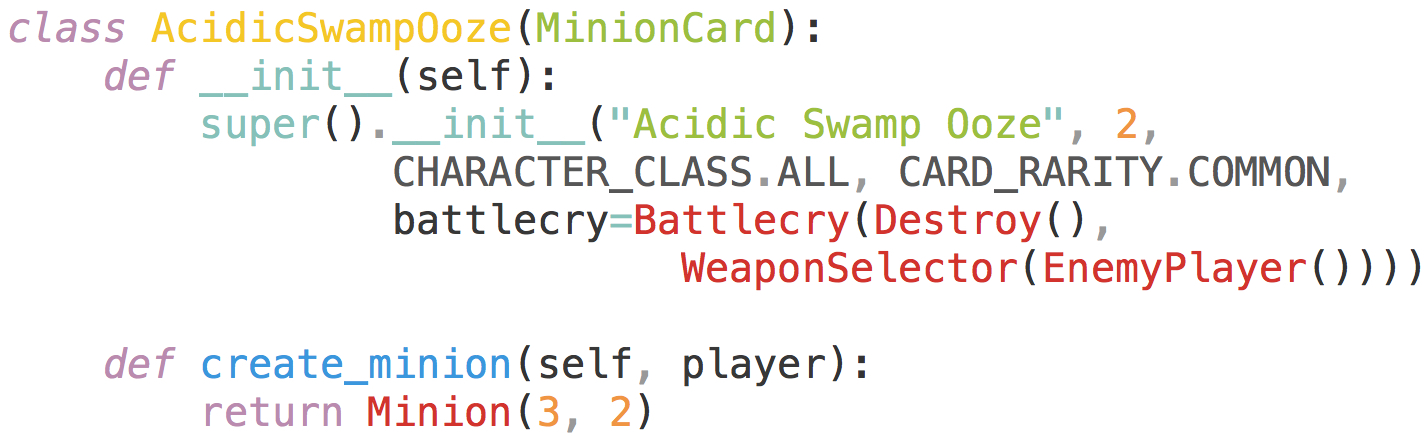}

		\caption{Example card of HearthStone. (a) Input description; (b) Output program.}
		\label{fig:hearth}
	\end{figure}

	Our model is trained by cross-entropy loss against the groundtruth program. Since our entire model is differentiable, all parameters are learned by gradient-based update. 
	
	For inference, we seek a sequence of grammar rules that maximizes the probability conditioned on input. The recursive prediction of the rules terminates if every leaf node in the (partial) tree is a terminal symbol. We use beam search to approximate the global inference, and the beam size is 5 in our experiments. Invalid rules for a particular node type are not considered during inference. For example, $\mathrm p_2\rightarrow\mathrm  c_1\mathrm c_2$ cannot be applied to the node $\mathrm p_1$ if $\mathrm p_1\ne\mathrm  p_2$.

	\section{Evaluation}
	In this section, we present experimental results of our CNN-based code generation. We evaluated our method on two types of tasks: (1) Python code generation for the HearthStone game, and (2) executable logic form generation for semantic parsing. 
    \subsection{Experiment \uppercase\expandafter{\romannumeral1}: HearthStone Code Generation}

	\subsubsection{Dataset.}		Our first (and main) experiment is based on an established benchmark dataset, HearthStone \cite[HS]{ling2016latent}. The dataset comprises 665 different cards of the HearthStone game; the input of each data point is a semi-structured description of fields, such as the card name, cost, attack, description, and other attributes; and the output is a Python code snippet that implements the functionality of the card, shown in Figure~\ref{fig:hearth}. We follow the train-dev-test split as in \newcite{ling2016latent}. The column HS in Table~\ref{tab:stat} lists relevant statistics of the dataset.
	\subsubsection{Metrics.}
	We evaluated our approach by accuracy and BLEU scores. Ideally, the accuracy should count the fraction of functionally correct programs, which unfortunately is not Turing computable. We followed most previous studies~\cite{ling2016latent,yin2017syntactic}, and calculated the accuracy based on string match (denoted as \textbf{StrAcc}).\footnote{Since spaces and empty lines are not represented in AST, the string match is computed based on a ``normalized'' format. Notice that indentation is crucial to a python program, which has been implicitly captured by AST.} We also find that several generated programs use a different variable name but implements a correct functionality, and that sometimes an argument name in a function call is or is not specified. Although different from the reference program, they are obviously correct programs after manual inspection, and we denote human-adjusted accuracy by \textbf{Acc+}. Here, we did not perform checking for non-obvious alternative implementation of an algorithm, and thus Acc+ is still a lower bound of functional accuracy.
    
The quality of the generated code is further evaluated by the \textbf{BLEU} score as an auxiliary metric, which computes how close the generated code is to the groundtruth code in terms of $n$-grams.

		\begin{table}[!t]
		\centering
		\small
		\begin{tabular}{llll}
			\toprule
			& \multicolumn{3}{c}{\textbf{Dataset}}\\
			\cmidrule{2-4}
			\textbf{Statistics} & \textbf{HS}  & {\textbf{ATIS}}  & {\textbf{JOBS}}\\
			\midrule
			\# Train & 533  & 4,434  &  500\\
			\# Dev & 66 & 491 &   -\\
			\# Test & 66 & 448  & 140\\
			\midrule
			Avg. tokens in description & 35.0 & 10.6 & 8.7\\ 
			Max. tokens in description & 76.0 & 48 & 22 \\
			Avg. tokens in code & 83.2 & 33.9 & 18.1\\
			Max. tokens in code & 403 & 113  & 50 \\
			Avg. nodes in AST & 151.0 & 47.2 & 40.0\\
			Max. nodes in AST & 744 & 154 & 138\\
			\bottomrule
		\end{tabular}
		\caption{Statistics of the datasets.}
		\label{tab:stat}
	\end{table}

	\begin{table}[!t]
		\centering
        \resizebox{\linewidth}{!}{
		\begin{tabular}{lrrr}
			\toprule
			\textbf{Model}  &\!\!\!\!\!\!\!\! \textbf{StrAcc} & {\textbf{Acc+}} &\!\!\! {\textbf{BLEU}} \\
			\toprule
			LPN~\cite{ling2016latent}  & 6.1 &--& 67.1 \\
            SEQ2TREE~\cite{dong2016language} &1.5&-- &53.4\\
			SNM~\cite{yin2017syntactic}& 16.2 &\!\!\!{\tiny$\sim$}18.2& 75.8 \\
			ASN~\cite{rabinovich2017abstract}& 18.2& -- & 77.6 \\
			ASN+SUPATT& \multirow{2}{*}{22.7}& \multirow{2}{*}{--} & \multirow{2}{*}{79.2}\\
            \quad\quad\cite{rabinovich2017abstract}&&&\\
			\midrule
			Our system &\textbf{27.3} & \textbf{30.3} & \textbf{79.6}\\
			\bottomrule		
		\end{tabular}}
		\caption{Performance of our model in comparison with previous state-of-the-art results. Accuracies are in percentage. \protect\newcite{yin2017syntactic} report an approximately 2\% percent boost after human adjustment.}\label{tab:HS}
	\end{table}
	
		\subsubsection{Settings.} 	For the input descriptions, we replace all punctuations with  a space; all letters are lower cased. For the neural network, we set the number of CNN layers $L$ to 21, where the bottom layer does not have skipping connections.  We also find it helpful to build a separate network (i.e., the same architecture, but with different weight parameters) for different AST node types (namely, nonterminal nodes, variable nodes, and function name nodes). This enables us to be better aware of node types during generation. When  predicting variable nodes, we introduce a new softmax target, for each slot, that could copy the slot value. The layers of difference CNN modules are set to the same dimension, chosen by validation from \{128, 192, 256\} for each predictor network. We applied dropout (drop rate$=0.5$) and $\ell_2$ penalty to regularize the fully connected layers.
	The network is trained by the Adam optimizer~\cite{kingma2014adam} with default hyperparameters. 
    
    \begin{table}[!t]
		\centering
        \resizebox{.9\linewidth}{!}{
		\begin{tabular}{rlrr}
			\toprule
			\textbf{Line \#}&\textbf{Model Variant}  & {\textbf{Acc+}} & {\textbf{BLEU}}\\
			\midrule
			1&        	Full model & \textbf{30.3} & 79.6\\
			\midrule
			2&			\  Pre-order CNN $\rightarrow$ LSTM  & 21.2 &78.8\\
			3&		    $-$ Predicted rule CNN & 24.2 & 79.2\\
			4&			$-$ Pre-order CNN & 25.8 &\textbf{80.4}\\
			5&			$-$ Tree-based CNN & 25.8 & 79.4\\
			6&			$-$ Tree-path CNN &28.8 &\textbf{80.4}\\
			7&			$-$ Attentive pooling & 24.2 & 79.3\\
			8&			$-$ Scope name & 25.8 & 78.6 \\
			\bottomrule
		\end{tabular}}
		\caption{Ablation test.}
		\label{tab:ablation}
	\end{table}

	\subsubsection{Overall Results.}
	Table~\ref{tab:HS} presents the results of our CNN-based code generation, in comparison with previous state-of-the-art models: (1) Latent Predictor Network~\cite[LPN]{ling2016latent}, an enhanced sequence-to-sequence model with multiple token-level predictors; (2) SEQ2TREE~\cite{dong2016language}, a sequence-to-sequence model based on AST; (3) Syntactic Neural Model~\cite[SNM]{yin2017syntactic}, an LSTM decoder based on AST structures; and (4) Abstract Syntax Networks~\cite[ASN]{rabinovich2017abstract}, another AST-based sequence-to-sequence model, which builds two LSTMs predicting rules in the horizontal and vertical directions, respectively. The ASN model has a variant (ASN+SUPATT) where attention is trained in a supervised fashion.
	
	As shown, our model outperforms all previous results in terms of both accuracy and BLEU scores. In particular, our accuracy is significantly higher than the previous state-of-the-art result by about 5 percentage points in terms of string accuracy. For human adjusted accuracy (Acc+), \newcite{yin2017syntactic} report approximately 2 percentage points improvement. Similar phenomena are observed in our scenario, and we achieve an Acc+ score of 30.3\%,
    showing strong evidence of the effectiveness of our approach. 
	
\begin{figure}[!t]
    \footnotesize
    \centering
    \includegraphics[width=\linewidth]{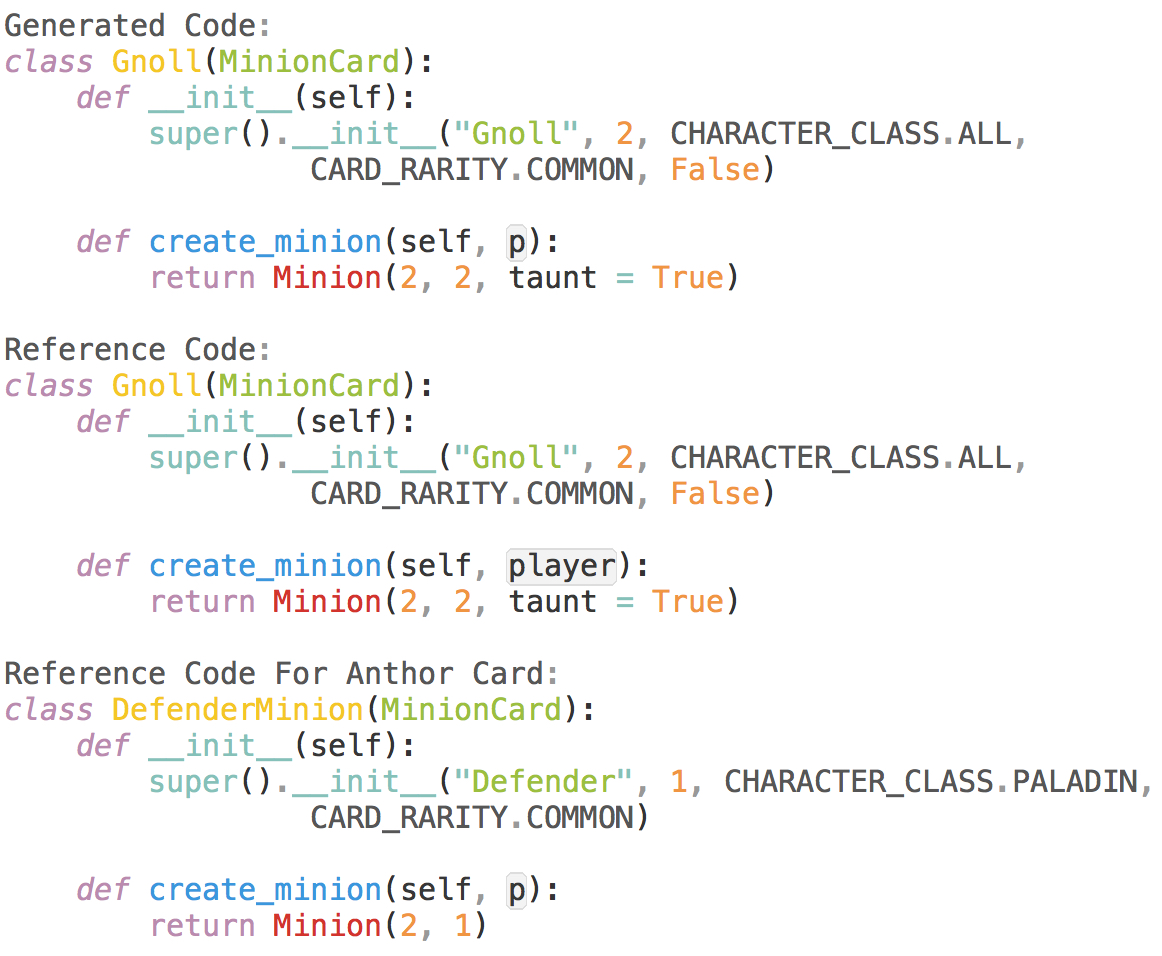}
    \caption{Example of the generated code. (Compared with reference codes, the code we generated has a different variable, but a correct functionality.)}
    \label{fig:bleu}
\end{figure}

	We find an intriguing fact that several previous methods could achieve a similar BLEU score to our approach, but with a much lower accuracy. For example, the ASN model has a BLEU score of 79.2, comparable to 79.6 given by our model. However, ASN only achieves 22.7\% string accuracy, whereas ours is 27.3\%. This is because the BLEU metric only measures surface $n$-gram similarity of programs. Previous methods (like ASN) are able to generate seemingly plausible code, which are in fact incorrect in terms of details. Therefore, we only consider BLEU scores (adopted in previous work) as a secondary metric. The major metric, namely, accuracy, shows that our approach generates much more accurate programs than previous models.

	\subsubsection{Ablation test.}	We conducted extensive ablation tests to analyze the contribution of each component. Although the development of our network started from a simple baseline and we incrementally added on useful components, the ablation test was conducted in an opposite way: it started from the full model, and we either removed a single component of our model or substituted it with some reasonable alternatives. We report results of our ablation test in Table~\ref{tab:ablation}.

	We first analyze the effect of CNN by substituting a CNN component with LSTM-based RNN (Lines~1 \&~2). Since the main information lies in the partial AST (151 nodes on average shown in Table~\ref{tab:stat}) as opposed to, say, the predicted grammar rules, we replace only the pre-order traversal CNN to an LSTM in this controlled experiment. Such setting achieves 1 point lower BLEU, but 9 percent lower accuracy. The result is  consistent with previous models in the literature (Table~\ref{tab:HS}), where RNN is the main building block achieving lower accuracy.
	
	The scenario can be better understood if we consider the setting where we simply remove the pre-order traversal CNN (Line 4, Table~\ref{tab:ablation}). Both based on a tree-based CNN layer, the LSTM component yields a 4\% lower  accuracy than simply without LSTM. 
	This implies that RNNs are not suitable to this task, which is probably because a program contains too many tokens and AST nodes. An RNN applied to such a long sequence can be very difficult to train{~\cite{pascanu2013difficulty}}, achieving significantly worse performance.
	
	We also analyze other components of our model, including the CNN for predicted rules, the tree-based convolutional layer, the tree-path convolution, the attentive pooling mechanism, and the scope controllers for pooling (Lines 3--8, Table~\ref{tab:ablation}). We see that each of the above components contributes to the whole model in its own way, improving the accuracy by 3--6 percent. These results show that we have designed reasonable components of the neural architecture, suited to the code generation task.

	\subsection{Experiment \uppercase\expandafter{\romannumeral2}: Semantic Parsing}
	\subsubsection{Dataset and Settings.}
	\begin{figure}[!t]
    \footnotesize
		\textbf{Input description:}\quad list airport in ci0     \\
        \textbf{Output $\lambda$-calculus:}
        
\quad\quad\quad\quad\includegraphics[width=.3\textwidth]{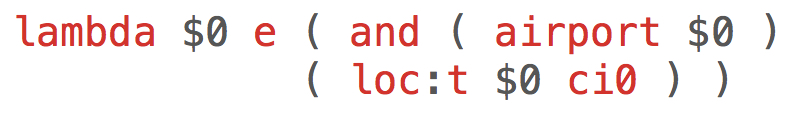}
		\caption{Example of the ATIS dataset for semantic parsing.}
		\label{fig:atis}
	\end{figure}
   Semantic parsing aims to generate logical forms given a natural language description. It can be thought of as code generation for a domain-specific language, because the logical form is an executable, unambiguous formal language. However, the style of semantic parsing differs significantly from python code generation. Since our model is mainly developed on the HS dataset, this experiment serves as additional evaluation of the generalizability of our model.
   
   We evaluated our model on two semantic parsing datasets (ATIS and JOBS) used in~\newcite{dong2016language}, where the input is a natural language sentence. The output of  ATIS is in the $\lambda$-calculus form (illustrated in Figure~\ref{fig:atis}), while for JOBS, it is in the Prolog-style form. We used the standard train-dev-test split for the datasets~\newcite{zettlemoyer2012learning}. 
 We see from the statistics in Table~\ref{tab:stat}  that the logical forms for semantic parsing contain significantly fewer nodes and tokens than HS Python code.
   
We adopted mostly the same network from the HearthStone experiment to semantic parsing. The number of layers $L$ is 7 in this experiment. We did not build a separate network for different node types as our network is prone to overfitting for such small datasets. Besides, we introduced the pointer network~\cite{see2017get} to copy variable names (e.g., \texttt{ci0} in Figure~\ref{fig:atis}) due to the property of the datasets, which is also the practice of previous work~\cite{rabinovich2017abstract}.

    \begin{table}[!t]
		\centering
        \resizebox{.9\linewidth}{!}{
		\begin{tabular}{l|lr|lr}
			\toprule
			& \multicolumn{2}{c|}{\textbf{ATIS}} & \multicolumn{2}{|c}{\textbf{JOBS}}\\
                        			\midrule
	\multirow{3}{*}{\rotatebox{90}{Traditional}}			& {\textbf{System}}  & {\textbf{Accuracy}} &{\textbf{System}}  & {\textbf{Accuracy}} \\ 
 
		&ZH15 & 84.2 &ZH15&85.0\\
			& ZC07&84.6&PEK03&88.0\\
			&WKZ14&\textbf{91.3}&LJK13&90.7\\
			\midrule
	\multirow{3}{*}{\rotatebox{90}{Neural}}		&SEQ2TREE&84.6&SEQ2TREE&90.0\\
			&ASN&85.3&ASN&91.4\\
			&ASN-SUPATT&85.9&ASN-SUPATT&\textbf{92.9}\\
            \midrule
			&Our System & 85.0 &Our System & 89.3\\
			\bottomrule
		\end{tabular}}
		\caption{Accuracy in semantic parsing (in percentage).}
		\label{table:sp}
	\end{table}

	\subsubsection{Results.}
		We followed \newcite{dong2016language} and evaluated our approaches by accuracy. It counts the fraction of exact match, except that we adjusted the order of conjunction and disjunction clauses to avoid spurious errors, as in all previous work.\yxmodify{}{ We did not measure BLEU since it is not used in existing studies.}

Table~\ref{table:sp} shows the performance of our model. As seen, neural models are generally worse than the WKZ14 system~\cite{wang2014morpho}, which uses a large number of rules and templates, but they outperform other traditional semantic parsing systems, including ZH15~\cite{zhao2015type}, ZC07~\cite{zettlemoyer2007online}, PEK03~\cite{popescu2003towards}, and LJK13~\cite{liang2013learning}.

We also see that our grammar-based structural CNN decoder achieves similar results to the state-of-the-art neural models~\cite{dong2016language,rabinovich2017abstract}. 
It should also be pointed out that in semantic parsing, we do not achieve a large performance boost as in HearthStone (HS) code generation. This is probably because the logic form for semantic parsing is usually short, containing only 1/4--1/3 tokens as in HS, and thus, both RNN and CNN are fine for logic form generation. This experiment nevertheless provides additional evidence of the generalizability and flexibility of our CNN code generation, since our model is basically designed for long programs (such as HS) but also works fine with semantic parsing.

	\section{Related Work}
	Early studies on code generation mostly focus on  domain specific languages~\cite{zettlemoyer2012learning,kushman2013using,wang2014morpho}. They are largely based on rules and human defined features, and thus are highly restricted.
	
Recently, researchers introduce neural networks to generate code in a general-purpose programming language. \newcite{ling2016latent} adopt a sequence-to-sequence model, but enhance it with multiple predictors. Other studies generate programs along abstract syntax trees~\cite{dong2016language,rabinovich2017abstract,yin2017syntactic}. However, their decoders are all based on RNNs, which are shown improper for code generation in our experiments.
	
	CNNs are origianlly used in classification tasks~\cite{lecun1995convolutional,krizhevsky2012imagenet}. \newcite{mou2016convolutional} propose a tree-based CNN to capture structural information. Such idea can be extended to general graphs, e.g., molecule analysis~\cite{duvenaud2015convolutional}. Recently, researchers develop deep CNNs for decoders~\cite{gehring2017convolutional,chaturvedi2018cnn}. In our paper, we incorporate the idea of structure-sensitive CNN and CNN for generation, and design a grammar-based structural CNN for code generation.

	\section{Conclusion}
	In this paper, we propose a grammar-based structural CNN for code generation. Our model makes use of the abstract syntax tree (AST) of a program, and generates code by predicting the grammar rules. We address the problem that traditional RNN-based approaches may not be suitable to program generation, possibly due to the large number of tokens/nodes in a program.  We thus design a CNN encoder-decoder model based on AST structures. 
    
    Our main experiment on the HearthStone dataset shows that we have achieved significantly better performance than previous RNN-based methods. Additional experiments on two semantic parsing tasks demonstrate the robustness of our approach. We also conducted in-depth ablation test to verify the effectiveness of each component in our model.

\section{Acknowledgments}
This work is sponsored by the National Key Research and Development Program of China under Grant No.~2017YFB1001803, and National Natural Science Foundation of China under Grant
Nos.~61672045 and 61529201.
	 
    \fontsize{9.0pt}{10.0pt}
	\bibliography{aaai}
	\bibliographystyle{aaai}
\end{document}